\documentclass[letterpaper, 10 pt, conference]{ieeeconf}  

\IEEEoverridecommandlockouts                              

\usepackage{graphics} 
\usepackage{xcolor}
\usepackage{adjustbox}
\usepackage{siunitx}
\usepackage{graphicx}
\usepackage{subcaption}
\usepackage{amsmath}
\usepackage{amssymb}
\usepackage{booktabs}
\usepackage{mathtools}
\usepackage{tabularx}
\usepackage{dsfont}
\usepackage{graphicx}
\usepackage{comment}
\usepackage{acronym}
\usepackage{multirow}
\usepackage{nicefrac}
\usepackage{times}
\usepackage{graphicx}
\usepackage{amsmath}
\usepackage{tabularx}
\usepackage{amssymb}
\usepackage{makecell}  
\usepackage{float}
\usepackage{booktabs}
\usepackage{afterpage}
\usepackage{diagbox}
\usepackage{colortbl}
\usepackage{booktabs}
\usepackage{threeparttable}
\usepackage{makecell}
\usepackage{cuted}
\usepackage{soul}
\usepackage{cite}
\usepackage{algorithm}
\usepackage{algpseudocode}
\usepackage{epsfig} 

\title{\LARGE \bf
GRITS: A Spillage-Aware Guided Diffusion Policy for \\ Robot Food Scooping Tasks
}
\author{%
\makebox[\linewidth][c]{%
Yen-Ling Tai$^{1*}$ \quad
Yi-Ru Yang$^{1*}$ \quad
Kuan-Ting Yu$^{2}$ \quad
Yu-Wei Chao$^{3}$ \quad
Yi-Ting Chen$^{1\dagger}$
}\\[0.3em]
\makebox[\linewidth][c]{%
$^{1}$National Yang Ming Chiao Tung University \quad
$^{2}$XYZ Robotics \quad
$^{3}$NVIDIA
}\\[-20em]
\thanks{%
$^{*}$Equal contribution. 
$^{\dagger}$Corresponding author.
}
}

\begin{document}

\maketitle
\thispagestyle{empty}
\pagestyle{empty}

\begin{strip}
\centering
\includegraphics[width=0.95\linewidth]{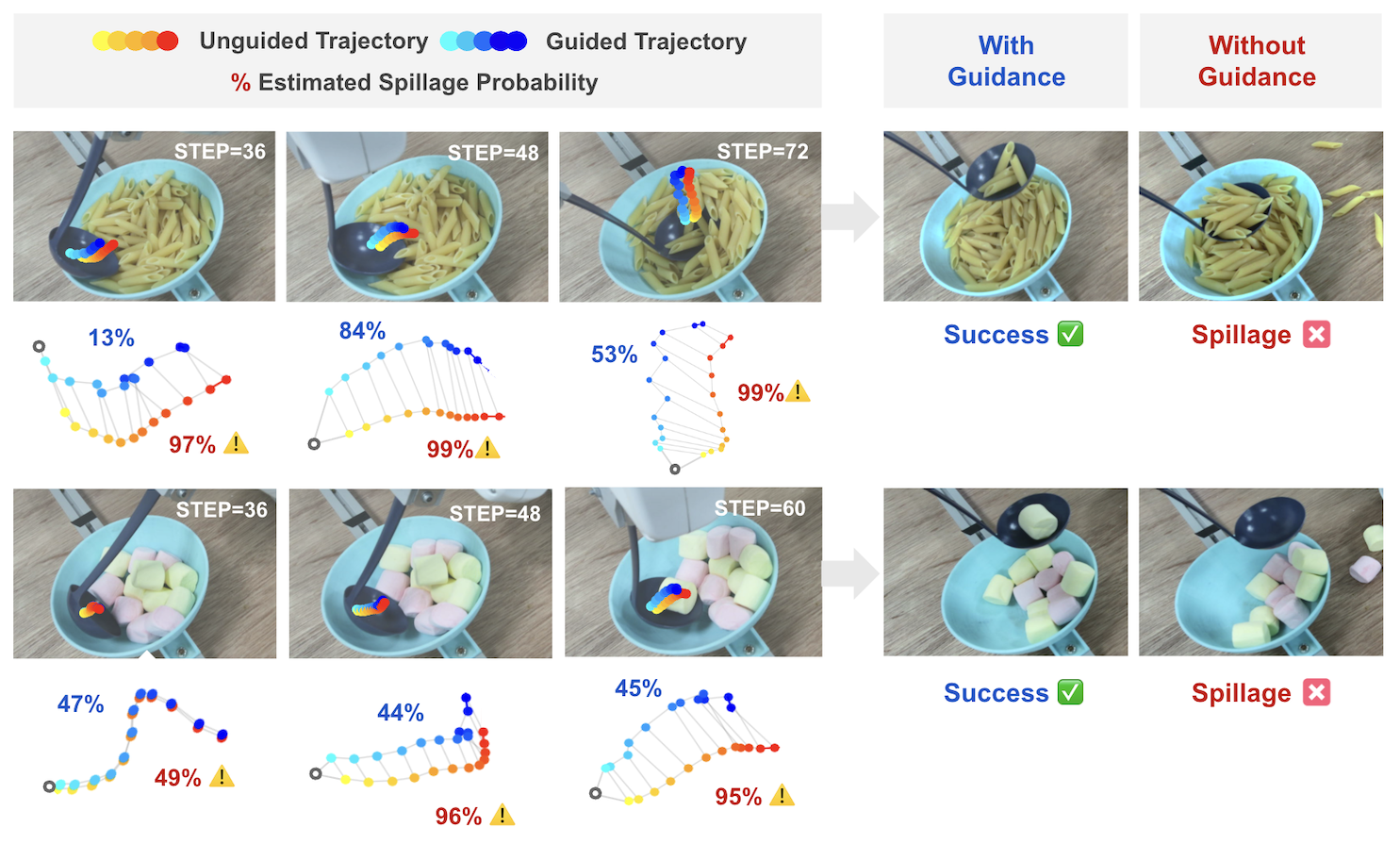}
\captionof{figure}{\textbf{Spillage-aware trajectory generation with GRITS.}
Robotic food scooping demands exact and delicate control, as small deviations can result in spillage.
GRITS addresses this challenge by leveraging predicted spillage probabilities to adaptively refine trajectories,
leading to safer and more reliable manipulation. The adjustments are subtle, with an average displacement of only
0.3 cm between consecutive trajectory points.}
\label{fig:teaser}
\end{strip}


\begin{abstract}

Robotic food scooping is a critical manipulation skill for food preparation and service robots. 
%
However, existing robot learning algorithms, especially learn-from-demonstration methods, still struggle to handle diverse and dynamic food states, which often results in spillage and reduced reliability.
In this work, we introduce GRITS: A Spillage-Aware \underline{G}uided Diffusion Policy for \underline{R}obot\underline{I}c Food Scoop \underline{T}ask\underline{S}. 
This framework leverages guided diffusion policy to minimize food spillage during scooping and to ensure reliable transfer of food items from the initial to the target location. 
%
Specifically, we design a spillage predictor that estimates the probability of spillage given current observation and action rollout.
The predictor is trained on a simulated dataset with food spillage scenarios, constructed from four primitive shapes (spheres, cubes, cones, and cylinders) with varied physical properties such as mass, friction, and particle size.
At inference time, the predictor serves as a differentiable guidance signal, steering the diffusion sampling process toward safer trajectories while preserving task success.
We validate GRITS on a real-world robotic food scooping platform. GRITS is trained on six food categories and evaluated on ten unseen categories with different shapes and quantities. 
GRITS achieves an 82\% task success rate and a 4\% spillage rate, reducing spillage by over 40\% compared to baselines without guidance, thereby demonstrating its effectiveness.
More details are available on our project website: \textcolor{magenta}{https://hcis-lab.github.io/GRITS/}.

\end{abstract}

\section{INTRODUCTION}

Robotic food scooping is a critical manipulation skill with broad applications in food preparation~\cite{singletary2022safety, xu2023roboninja, ye2024morpheus, zhang2019leveraging} and assistive feeding~\cite{bhaskar2024lava, feng2019robot, grannen2022learning, jenamani2024flair, keely2025kiri, liu2024adaptive, liu2024imrl, sundaresan2022learning, sundaresan2023learning, tai2023scone}. 
%
%
These technological advances can boost efficiency, ease labor shortages, and improve quality of life for care recipients and caregivers.
%
%
%
Recently, learning from demonstrations (LfD) has attracted great attention 
for robot food scooping~\cite{bhaskar2024lava, liu2024adaptive, liu2024imrl, tai2023scone}.
%
%
These methods have shown great promise in learning policies for manipulating dynamic and deformable food items.
However, these methods are limited by the coverage of available demonstrations because collecting data across diverse food types and conditions is costly and labor-intensive.
This constraint prevents methods from generalizing to unseen scenarios and adapting to evolving deployment requirements. 
%

Among emerging approaches, diffusion policies stand out for their strong capability and generalization across diverse robotic manipulation tasks, including food manipulation~\cite{chi2023diffusion, xue2025reactive}. 
%
%
%
%
These models are appealing as they demonstrate strong generalization and robustness with relatively few expert demonstrations. 
While these models can favorably imitate demonstrations, they fail to account for unseen food states such as a bowl full of food, as shown in Fig.~\ref{fig:teaser}, and that often causes unintended spillage. 
Recently, guided diffusion policies have emerged. 
They enable test-time guidance that refines motion trajectories based on the intended objectives, while ensuring task success without retraining.
%
%
For example, the community has applied test-time guidance to 
steer diffusion models toward specified task outcomes~\cite{janner2022planning}, avoid undesirable states such as collisions~\cite{li2024language}, and ensure stability and smoothness of robot trajectory~\cite{carvalho2023motion}. 
%
%
%
This leads our central question: Can guided diffusion policy be leveraged to adjust trajectories in real-world food scooping, ensuring task success while avoiding risky situations such as spillage?
%

In this work, we introduce GRITS: A Spillage-Aware \underline{G}uided Diffusion Policy for \underline{R}obot\underline{I}c Food Scoop \underline{T}ask\underline{S}.
GRITS exploits diffusion policy~\cite{chi2023diffusion} and proposes a novel guidance mechanism to minimize food spillage during scooping and to ensure reliable transfer of food items from the initial to the target location.
%
%
Specifically, we design a spillage predictor that estimates the probability of spillage given a current observation and an action rollout.  
%
%
However, training such a predictor in the real world is impractical, as data collection would require tremendous effort to create spillage scenarios and manage cleanup. 
To overcome this challenge, we utilize Isaac Lab~\cite{mittal2023orbit}, a high-fidelity simulator, to construct a simulated food scooping dataset. 
We design four primitive shapes (spheres, cubes, cones, and cylinders) and assign them diverse physical properties such as mass, friction, and particle size, which are randomized within specified ranges during dataset collection. This design synthesizes a broad set of controllable food-like items and spillage scenarios. 
We choose point clouds as the predictor’s input representation to mitigate the sim-to-real gap. 
%
The trained spillage predictor is integrated into the denoising process of a guided diffusion policy to continuously evaluate spillage risk.
%
%
%
This contrasts with typical classifier-based diffusion models~\cite{dhariwal2021diffusion}, which rely on static labels during image generation.
%
%
%
%
%
To the best of our knowledge, this is the first work to apply guided diffusion policies to robotic food scooping, a task that requires dynamic rollout adjustment.
%
%
%
%
%
%

In our experiments, we demonstrate the effectiveness of GRITS in successfully scooping a wide range of food items while significantly reducing spillage. 
We compare GRITS against several strong baselines, including rule-based methods, learning-based approaches without spillage guidance~\cite{tai2023scone, chi2023diffusion}, and variations of guided diffusion policies. 
Experiments are conducted on a real-world robotic platform with diverse food types varying in shape, particle size, and quantity, with \textbf{six} granular food categories for training and \textbf{ten} unseen categories for testing.
The results show that GRITS generalizes well to novel scenarios, achieving an 82\% success rate and the lowest spillage rate of 4\%, which represents a reduction of more than 40\% compared to standard diffusion policy baselines.

In summary, our main contributions are as follows:
\begin{itemize}

    \item 
    We present GRITS, a novel spillage-aware guided diffusion policy that refines trajectories at the test time through continuous spillage-risk estimation for robotic food scooping.
    

    \item We introduce a simulated data collection pipeline for scooping diverse food shapes with varying physical properties. 
    %
    
    \item We demonstrate that GRITS effectively scoops diverse and unseen food items and significantly reduces spillage compared to strong baselines in real-world experiments. 
\end{itemize}


\begin{figure*}[!ht]
    \centering
    \captionsetup{type=figure}
    \includegraphics[width=0.98\linewidth]{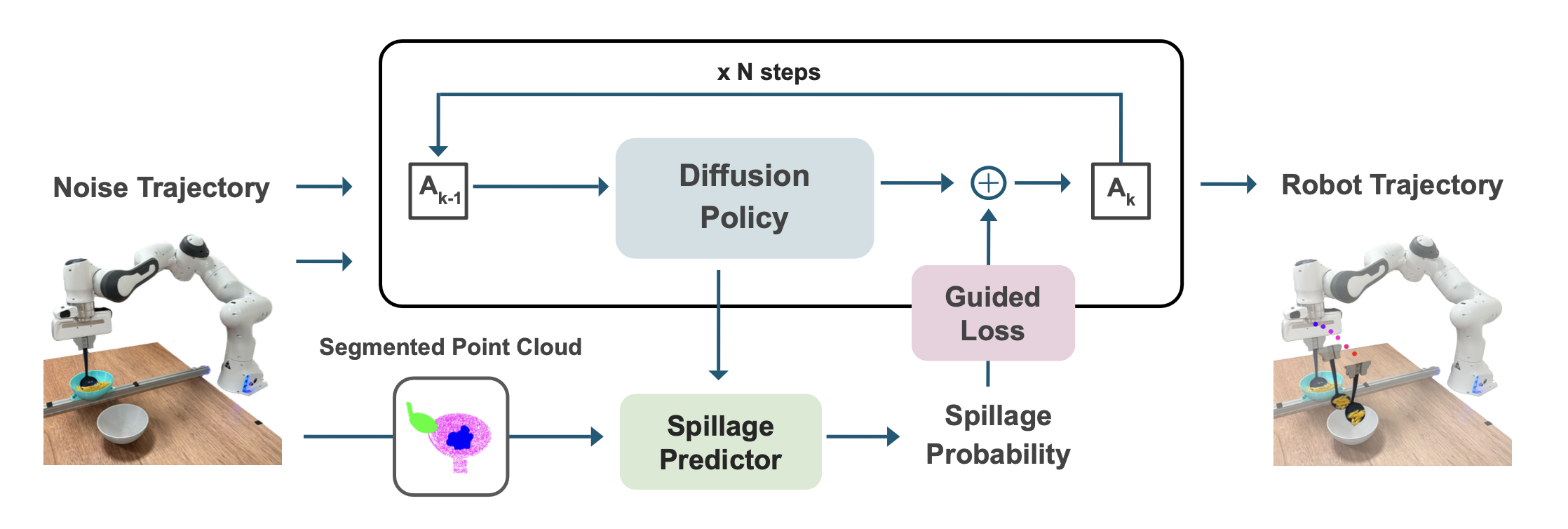}
    \caption{\textbf{The architecture of GRITS.} 
    GRITS is a guided diffusion policy designed for robotic food scooping. 
    Given an RGB-D image and an initial noisy trajectory, the diffusion policy denoises it into a refined trajectory.
    A spillage predictor, which takes segmented point clouds as input to reduce the sim-to-real gap, estimates the probability of spillage for given candidate trajectory.
    This probability provides a guidance signal that steers the denoising process toward safer trajectories. 
    The robot then follows the refined trajectory using position control to scoop food items.
    }
    \label{fig:GRITS}
\end{figure*}

\section{RELATED WORK}

\subsection{Robotic Food Manipulation}

Food manipulation has attracted significant attention in the robotics community, with increasing work on applications such as food preparation~\cite{singletary2022safety, xu2023roboninja, ye2024morpheus, zhang2019leveraging} and assistive feeding~\cite{bhaskar2024lava, feng2019robot, grannen2022learning, jenamani2024flair, keely2025kiri, liu2024adaptive, liu2024imrl, sundaresan2022learning, sundaresan2023learning, tai2023scone}. 
%
%
Within this domain, robotic food scooping has emerged as a core task, driving research on both system design and algorithms.
Recent advances focus on addressing specific challenges: developing learning frameworks to handle diverse food properties and types~\cite{bhaskar2024lava, grannen2022learning, liu2024adaptive, liu2024imrl, tai2023scone}, designing specialized hardware for food acquisition~\cite{keely2025kiri}, and creating integrated assistive feeding systems~\cite{park2020active}.
These learning-based approaches are instances of learning from demonstrations (LfD), relying on high-quality expert demonstrations~\cite{argall2009survey, atkeson1997robot}.
%
%
However, collecting demonstrations across diverse foods and conditions is costly, and existing approaches mainly optimize for task success without accounting for dynamic food states and fine-grained control, often leading to spillage in real-world scenarios.
%
In this work, we present a new LfD framework based on guided diffusion policies that accounts for unseen food states, preventing spillage while ensuring successful scooping. 
%

\subsection{Guidance Mechanism for Diffusion Models}

Diffusion models are a powerful class of generative models that have gained significant attention for image and video generation~\cite{ho2020denoising, nichol2021improved, sohl2015deep, song2020score}.
These models employ a chain of incremental updates in both forward and backward processes to learn the underlying multimodal data distribution.
Recently, diffusion models have also shown strong potential in sequential decision-making for robotics~\cite{ajay2022conditional, chi2023diffusion, janner2022planning, reuss2023goal, wang2022diffusion, ze20243d}, particularly for handling multimodal action distributions.
They have been explored in reinforcement learning~\cite{ajay2022conditional, janner2022planning, wang2022diffusion} and imitation learning~\cite{chi2023diffusion, reuss2023goal, ze20243d} for robot manipulation.
More recently, the community has explored the power of guidance mechanisms that enable the controllability of diffusion models and test-time adaptation~\cite{carvalho2023motion, dhariwal2021diffusion,  huang2023diffusion, janner2022planning, lee2024learning, li2024language, saha2024edmp, xu2024dynamics}.
In robotic manipulation, guidance has been explored through three main objectives: goal-related objectives~\cite{carvalho2023motion, huang2023diffusion, janner2022planning, lee2024learning, xu2024dynamics}, which guide the agent to reach a target state; physics-related objectives~\cite{carvalho2023motion, huang2023diffusion, li2024language, saha2024edmp}, which ensure stability and smoothness of robot trajectory while minimizing safety risks like collision to surroundings; and rule-related objectives~\cite{janner2022planning}, which encourage adherence to task rules and success criteria. 
%
While these approaches have shown promise, there remains significant potential in extending guidance mechanisms to more complex real-world challenges.
%
%
In this work, we take a first step by introducing spillage-aware objectives into guided diffusion policies, enabling reliable rollouts for robotic food manipulation.
%
%
%


\begin{figure*}[t!]
    \centering
    \captionsetup{type=figure}
    \includegraphics[width=0.98\linewidth]{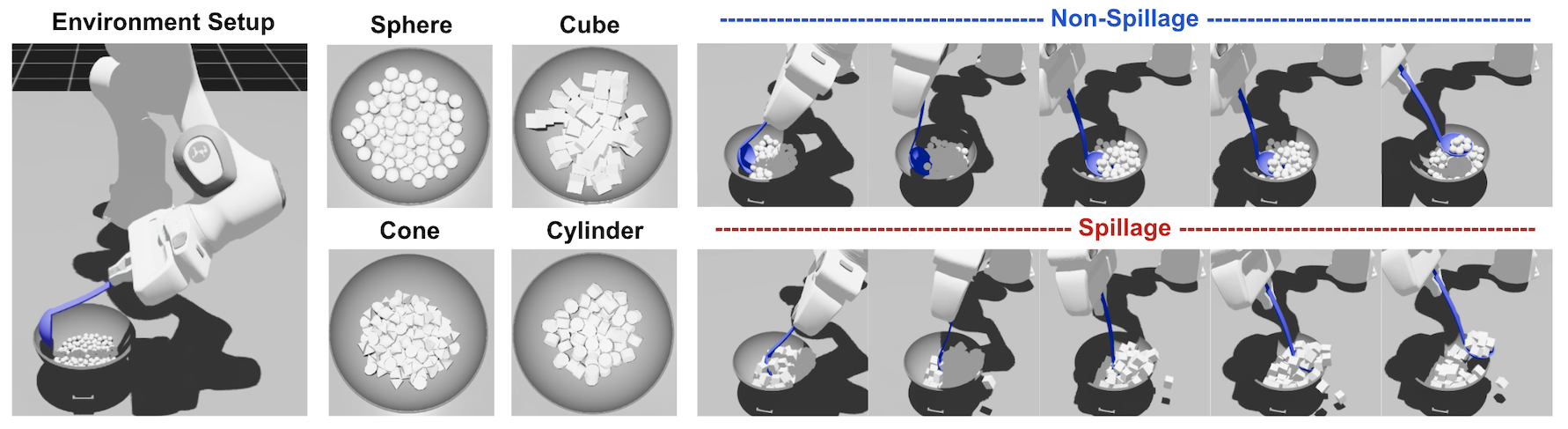}
    \caption{\textbf{Simulated Food Scooping Data Collection.} 
    We construct a scooping dataset in simulation to train the spillage predictor. 
    Simulated foods are composed of four primitive shapes: spheres, cubes, cones, and cylinders, with varied physical properties, including mass, friction, and particle size. 
    This design enables diverse and controllable scooping and spillage cases under different rollouts, which are impractical to collect in the real world.
    %
    %
    }
    \label{fig:spillage_predictor}
\end{figure*}

\section{METHOD}

\subsection{Problem Formulation}

In robotics food scooping, a robot is tasked to scoop food items from a specified container (fixed) and transfer them to the target container (randomly placed). 
This task can be divided into three stages: scooping, transferring, and pouring. Among these, scooping is the most challenging, as it involves complex interactions with diverse food properties and thus is formulated as a policy learning problem. 
We use bowl location detection with motion planning for transferring, and predefined trajectories for pouring.

Given a sequence of past and current \textit{m}-step RGB-D observations $O_{t}=\{v_{t-m+1}, …v_{t-1}, v_{t}\}$, a scooping policy outputs a sequence of the next \textit{n}-step robot end-effector poses $A_{t}=\{a_{t}, a_{t+1}, …a_{t+n-1}\}$.
A scoop is successful if it lifts at least one large item or covers one-third of the spoon with small-particle foods. 

\subsection{Guided Diffusion Policy}

We explore guided diffusion policy as a solution for handling diverse food properties and minimizing spillage in robotic scooping. 
Guided diffusion policy decouples training and inference to balance generalization and task-specific adaptation. 
During training, the diffusion policy learns to imitate expert trajectories without explicit task-oriented constraints, supporting broad generalization.
At inference, a spillage-aware objective refines trajectories to minimize spillage and ensure scooping success.
This design simplifies training and allows flexible adaptation to real-world conditions without retraining.

\subsubsection{Diffusion Policy Training} 
To model the conditional distribution $P(A_{t}|O_{t})$, Diffusion Policy~\cite{chi2023diffusion} formulates trajectory generation as a reverse denoising process. A noisy trajectory is first sampled from a Gaussian distribution, and a noise predictor $\epsilon_{\theta}$ iteratively estimates the noise at each step $k$, progressively refining the trajectory toward a clean action sequence. The reverse process is parameterized by a noise scheduler with hyperparameters $\sigma$, $\alpha$, and $\gamma$, which balance the capture of high- and low-frequency action characteristics. Training is performed by minimizing the mean squared error (MSE) between the predicted noise $\epsilon_{\theta}$ and the true noise $\epsilon$.
%

\vspace{-0.5em}
\begin{equation}
    A_{k-1}= \mu_k + \sigma_kz, \quad z\sim(0,I)
\end{equation}

\vspace{-1.5em}
\begin{equation}
    \mu_k = \alpha(A_k-\gamma\epsilon_{\theta}(O_t, A_t, k))
\end{equation}


\subsubsection{Guided Sampling} 
At inference, we incorporate a guidance mechanism to steer generated trajectories using an objective function $J$.
In our case, $J$ is defined by the spillage predictor, which outputs a continuous risk score given the current observation and candidate rollout.
Since the objective function is differentiable with respect to the rollout, we can compute $\nabla J$ and use it to refine trajectories by minimizing $J$.
%
At each denoising step $k$, we estimate the clean trajectory $A_{0|k}$, evaluate $J(A_{0|k}, P_t)$, compute its gradient, and update $A_{k-1}$ accordingly.
The guidance weight $\rho$ controls how strongly the objective influences trajectory generation.
%
%
%

\vspace{-1em}
\begin{equation}
    A_{0|k}:=\mathbb{E}[A_0|A_k]=\frac{A_k-\sqrt{1-\overline{\alpha}_k\epsilon_{\theta}(A_k)}}{\sqrt{\overline{\alpha}_k}}
\end{equation}

\vspace{-1.3em}
\begin{equation}
    A_{k-1}=\mu_k-\rho\nabla J + \sigma_kz
\end{equation}
\vspace{-1.4em}


\subsection{Spillage-Aware Guided Diffusion Policy}

\subsubsection{Simulated Food Scooping Data Collection}
The proposed spillage
predictor aims to estimate the probability of spillage given a current observation and an action rollout.
%
%
However, training it in the real world is impractical, as generating spillage scenarios requires heavy effort for setup and cleanup and risks damaging food items.
%
These factors motivate us to investigate a simulation-based solution.
%
%
%

We utilize Isaac Lab~\cite{mittal2023orbit} to construct an environment with hardware settings that closely replicate the real-world setup to mitigate the sim-to-real gap.
Fig.~\ref{fig:spillage_predictor} depicts our simulation development environment.
%
%
We construct a dataset of 4,000 scooping trials, with 2,000 spillage and 2,000 non-spillage.
Trajectories are diversified by applying random shifts to a mean trajectory derived from real-world demonstrations, with noise ranges of ±7 cm along the X-axis and ±5 cm along the Y and Z axes.
To ensure feasibility, the perturbations are constrained by both the base trajectory and bowl size.
Further diversity is introduced by varying simulated object properties, including particle shape, radius, mass, friction, and quantity.
%
%
%
The dataset spans four primitive shapes: spheres, cubes, cones, and cylinders, with 1,000 trials for each. 
%
 
\begin{figure*}[ht!]
    \centering
    \includegraphics[width=0.95\textwidth]{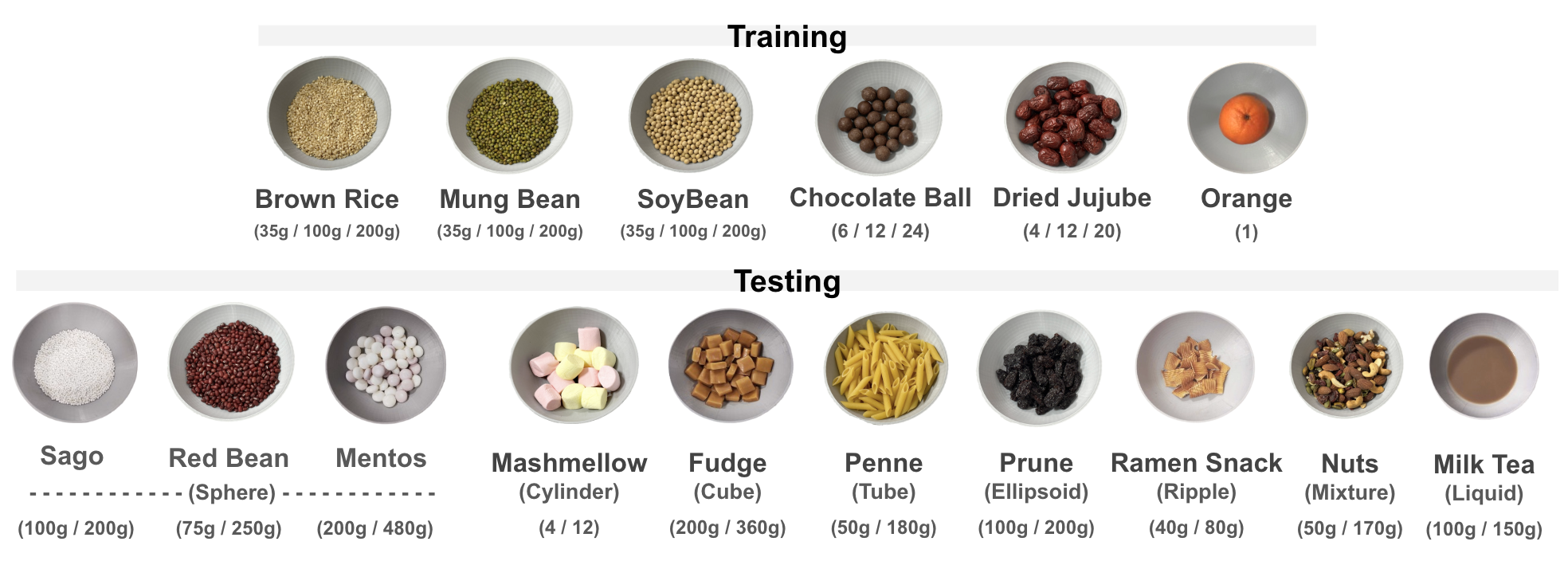}
    \caption{\textbf{Food categories for training and testing sets.} The training set includes six food items (top row) varying in sphere size from small to large. The testing set (bottom row) covers ten additional food categories with diverse shapes and material properties. Numbers below each food item indicate quantities: small-particle foods are measured by weight (g), and large-particle foods are measured by count (pieces).
    }
    \label{fig:food_category}
\end{figure*}

\subsubsection{Spillage Predictor Training} The spillage predictor $f_\mathtt{spillage}$ estimates the probability of spillage ${P_{\mathtt{spillage}}}$ by taking a generated trajectory ${A_t}$ and a segmented object point cloud ${P_t}$. 
We use segmented point clouds $P_t$ as the environment representation to reduce the sim-to-real gap. 
To further reduce the influence of irrelevant scene elements, we apply SAM2~\cite{ravi2024sam} to segment pixels corresponding to spoon, bowl, and food.
%
%
The food point cloud is reconstructed from depth and the spoon and bowl are represented by their known CAD models, aligned via forward kinematics (for the spoon, rigidly attached to the end-effector) and a predefined calibrated pose (for the fixed bowl), and merged to form the composite point cloud.
%
%
We downsample each object’s point cloud to 3,000 points, reducing computational cost while keeping consistent input dimension across object types.

We encode object point clouds using the DP3 encoder~\cite{ze20243d}, a variant of PointNet++~\cite{qi2017pointnet++} that omits the T-Net and BatchNorm layers, which is effective for control tasks with a fixed camera and stable viewpoint.
%
%
%
%
The predicted trajectory ${A_t}$ is encoded with a fully connected MLP, and its embedding is concatenated with the point cloud features.
This joint representation is passed through an MLP decoder to produce a probability distribution over spillage and non-spillage classes via softmax.
%
%
%
The model is trained with cross-entropy loss between the predicted probability ${P_\mathtt{spillage}}$ and the ground-truth label $y \in \{0, 1\}$.
%
%

\subsubsection{Guided Diffusion Policy with Spillage Predictor}  
In this work, we treat the objective function $J(A_{0|k},P_t) = f_{spillage}(A_{0|k},P_t)$.
Given the generated trajectory and the observed segmented point clouds, the objective is to minimize the spillage probability, which can also be interpreted as the "distance" to a safer state. 
%
%
The full procedure is presented in \textbf{Algorithm 1}.



\begin{algorithm}[ht!]
    \caption{Spillage-Aware Guided Diffusion Policy}
    \begin{algorithmic}[1]
    \Require diffusion model \( \epsilon_\theta \), spillage predictor \(f_\mathtt{spillage}\), guidance weight \( \rho \), threshold \(s\)
    \State \( A_T \gets \text{sample from } \mathcal{N}(0, I) \)
    \For{\( k = T \) to \( 1 \)}
        \State \( A_{k-1} \gets \mu_k + \sigma_k z \)
        \If{\( k \leq s \)}
            \State \( A_{0|k} \gets \frac{A_k - \sqrt{1-\alpha_k} \epsilon_\theta(A_k)}{\sqrt{\alpha_k}} \)
            \State \( A_{k-1} \gets A_{k-1} - \rho \nabla_{A_{k}} f_{spillage}(A_{0|k},P_t) \)
        \EndIf
    \EndFor
    \State \Return \( A_0 \)
    \end{algorithmic}
\end{algorithm}

We set a threshold $s{=}30$ to delay guidance activation, allowing the diffusion policy to form a coarse trajectory before refinement and preventing over-correction in early denoising steps.
We adopt DDIM~\cite{song2020denoising} following Diffusion Policy~\cite{chi2023diffusion}, with $T{=}100$ training steps and $T{=}16$ denoising steps for evaluation and guidance weight $\rho{=}2.5$.
These settings balance trajectory quality, inference efficiency, and refinement effectiveness.
%



\begin{figure}[!ht]
    \centering
    \includegraphics[width=0.45\textwidth]{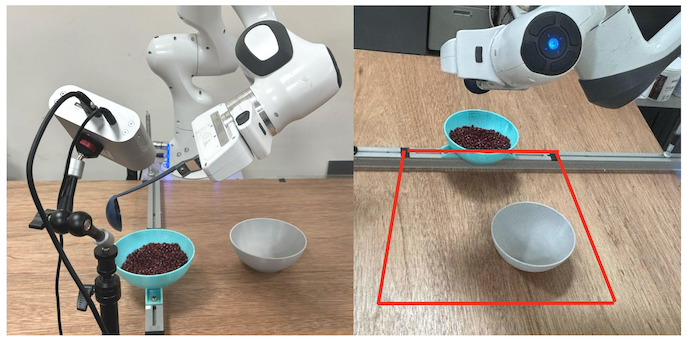}
    \caption{\textbf{Real-World Experimental Platform.} We set up a 35 × 30 cm workspace (indicated by the red bounding box) for the experiments.
    }
    \label{fig:real_world_setup}
    \vspace{-1.5em}
\end{figure}

\section{EXPERIMENTS}
\vspace{-0.5em}

\begin{figure*}[ht!]
    \centering
    \includegraphics[width=0.92\textwidth]{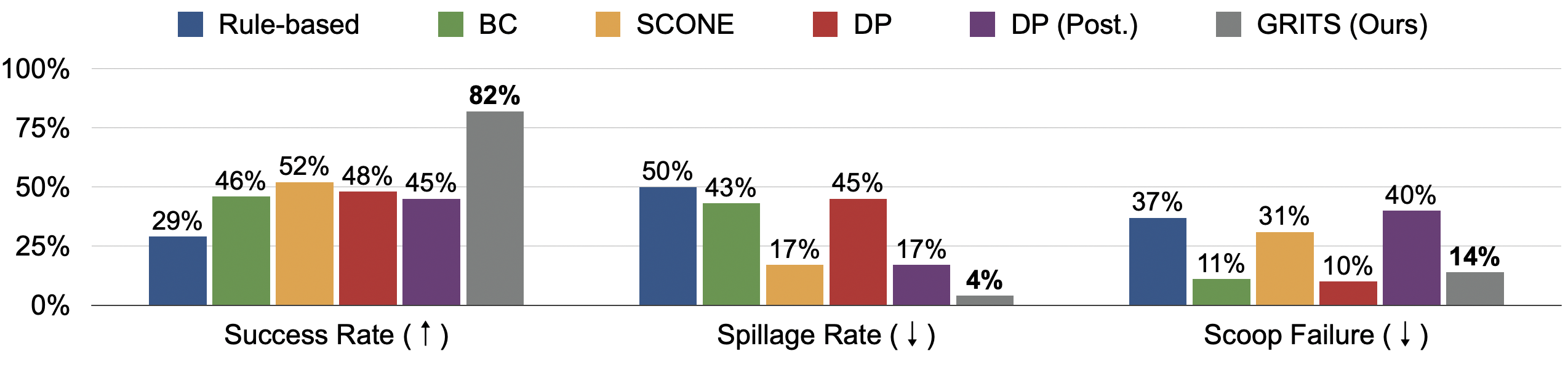}
    \caption{\textbf{Real-world experiment results.} GRITS achieves the highest task success rate 82\% while simultaneously maintaining the lowest spillage rate 4\% and a low scoop failure rate 14\%. 
    These results highlight the effectiveness of integrating spillage-aware guidance into diffusion models for robust and reliable food scooping.
    Notably, the scoop failure rate is slightly higher than vanilla DP, reflecting a trade-off where guidance reduces risky trajectories but may occasionally miss food collection.
    }
    \label{fig:result}
\end{figure*}

\begin{figure*}[!ht]
    \centering
    \includegraphics[width=0.92\textwidth]{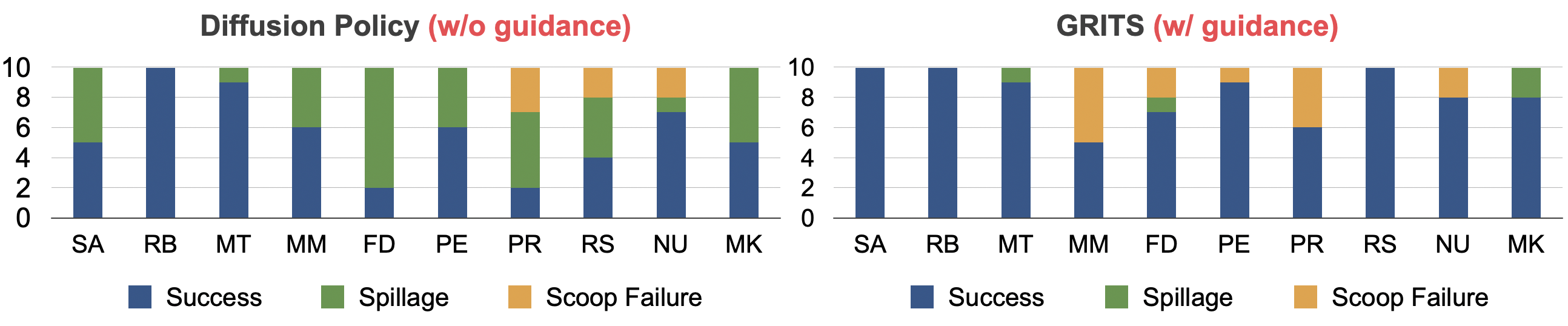}
    \caption{
    \textbf{Per-food comparison between vanilla diffusion policy (left) and GRITS (right).} The figure clearly highlights the effectiveness of GRITS in reducing spillage and ensuring scooping success.
    Unlike the previous figure, here we classify a trial as a spillage failure if spillage occurs at any point before a scoop is completed, regardless of whether food is eventually lifted, providing a more intuitive comparison.
    %
    %
    (SA: Sago, RB: Red Bean, MT: Mentos, MM: Marshmallow, FD: Fudge, PE: Penne, PR: Prune, RS: Ramen Snack, NU: nuts, MK: Milk Tea) 
    }
    \label{fig:detailed_result}
\end{figure*}

\subsection{Experiment Setup}

In the real-world setup, we use a 7-DoF Franka Emika Panda robot with a spoon attachment and two fixed Orbbec Femto Bolt RGB-D cameras, as shown in Fig.~\ref{fig:real_world_setup}. One container is fixed, and the other is randomly placed within a 35 $\times$ 30 cm workspace. The robot controller operates at 10 Hz.
To train the scooping policy, we collect expert demonstrations via record-and-replay procedure~\cite{tai2023scone}: a human kinesthetically guides the robot to record end-effector trajectories, which are then replayed to capture corresponding sensory data. 
The collected dataset includes six food categories: brown rice, soybeans, mung beans, chocolate balls, dried jujube, and orange. For each category, three quantity levels and five bowl configurations are used, except for orange, which has only one quantity level. A total of 80 demonstrations are collected.
For evaluation, we include ten additional food categories spanning sphere-like items and diverse shapes, each tested at two quantity levels with five trials per level. Fig.~\ref{fig:food_category} shows all the food categories and corresponding quantities.

\subsection{Baselines}


We compare against rule-based or naive cloning methods without a spillage predictor, and diffusion policies with alternative guidance mechanisms.

\begin{itemize}

\item \textbf{Rule-based:} 
The robot moves to the bowl center, measures food height using RGB-D images, then rotates to a predefined pose and digs to a fixed depth of 5 cm before executing a fixed scooping action. 

\item \textbf{Behavior Cloning (BC):} Naive policy learning that uses a CNN-based image feature extraction module. The extracted latent features are passed through an MLP to generate robot actions.

\item \textbf{SCONE~\cite{tai2023scone}:} A powerful scooping framework that integrates active perception to improve food understanding and generalization to unseen categories. It provides a strong baseline via perception-driven adaptability, but does not explicitly address safety issues such as spillage.

\item \textbf{Diffusion Policy (DP)~\cite{chi2023diffusion}:} A CNN-based diffusion policy without guidance, representing the core policy model used in GRITS.

\item \textbf{Diffusion Policy (Post.):} An extension of the vanilla diffusion policy where a spillage predictor evaluates trajectories after denoising and adjusts high-risk motions to reduce spillage. This enables spillage objectives without retraining, but since adjustments are applied post-hoc and do not consider task success, it is less effective than integrating guidance directly during denoising process.

\end{itemize}

\subsection{Evaluation Metrics}

We report three performance metrics evaluated over all trials, defined separately and potentially overlapping to capture complementary aspects of task performance:

\begin{itemize}

\item \textbf{Task Success Rate (\%):} The percentage of trials where food is successfully scooped and transferred without spillage, with success defined as either at least one item for large food pieces, or one-third of the spoon surface covered for small particle foods.

\item \textbf{Spillage Rate (\%):} The percentage of trials in which food is spilled during scooping.

\item \textbf{Scoop Failure Rate (\%):} The percentage of trials in which the robot fails to lift food. Notably, methods with frequent scoop failures may exhibit low spillage rates simply because nothing is scooped.

\end{itemize}

\begin{figure*}[!ht]
    \centering
    \includegraphics[width=0.94\textwidth]{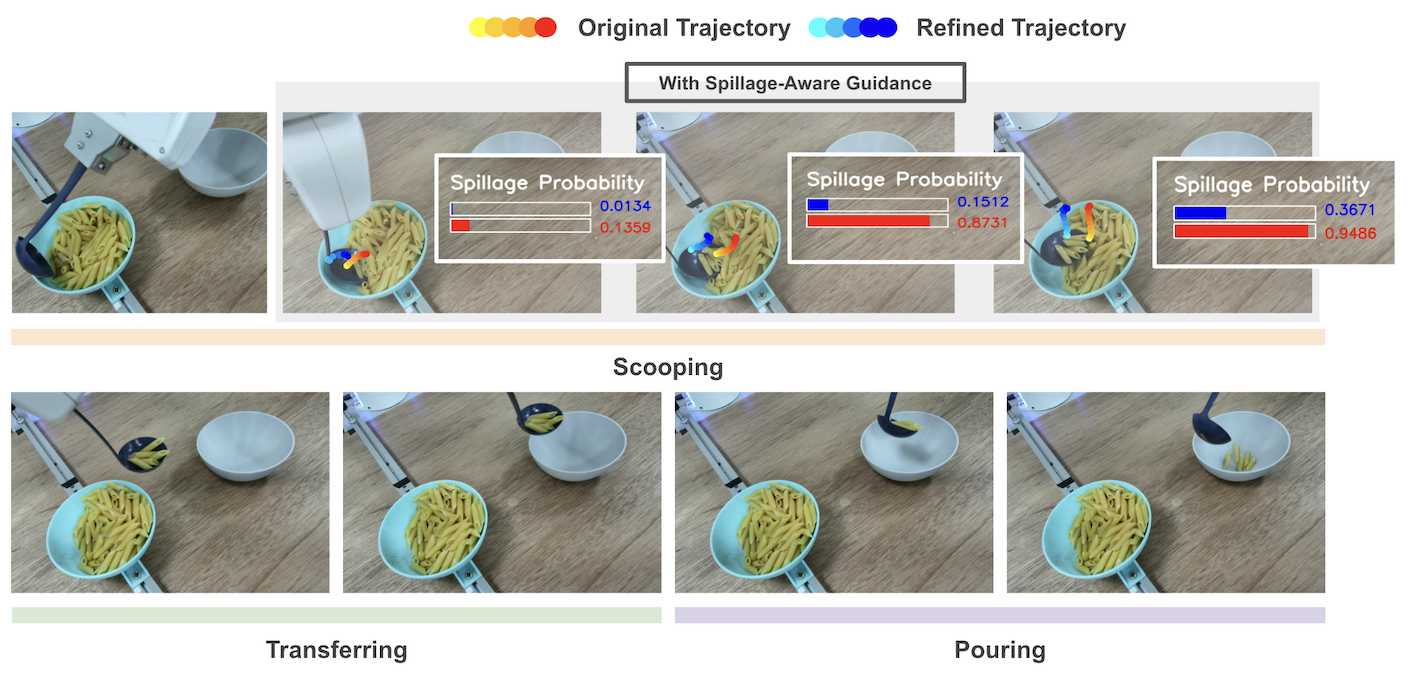}
    \caption{
    {
    \textbf{Real-world rollout of GRITS.} The figure shows GRITS scooping penne, where spillage-aware guidance refines the original trajectory into a safer one by reducing high-risk states, followed by transferring the food into the target container. 
    %
    %
    }
    }
    \label{fig:real_deployment}
\end{figure*}

\subsection{Real-World Experiments}

We evaluate GRITS on ten unseen food categories, spanning diverse shapes (spheres, cylinders, cubes, tubes, ellipsoids, ripples) and types (solids, mixtures, liquids). Results are summarized in Fig.~\ref{fig:result}. 
Overall, GRITS achieves the highest success rate of 82\% and the lowest spillage rate of 4\%, significantly outperforming all strong baselines.

%
Baselines without spillage predictor perform poorly. The rule-based method fails to adapt actions to different food items, underscoring the need for learning-based approaches.
%
%
Behavior cloning and vanilla diffusion policy improve success but still suffer from high spillage rates, raising safety concerns.
SCONE, which incorporates active perception, outperforms naive imitation learning by reducing spillage, but it suffers from higher scooping failure rates when handling very small food quantities, where interaction increases food state complexity.
Adding spillage-aware guidance reduces spillage across all baselines. 
The diffusion policy with post-processing achieves a lower spillage rate of 8\%. However, it suffers from a high scoop failure rate 40\% due to the lack of task-oriented guidance, resulting in low overall success.
Among all methods, GRITS achieves the best balance, demonstrating strong robustness across food categories.
We also observe a slightly higher scoop failure rate than the vanilla diffusion policy, reflecting a trade-off where spillage-aware guidance reduces risky trajectories but occasionally misses food collection.

Fig.~\ref{fig:detailed_result} compares the per-food performance of the vanilla diffusion policy and GRITS, showing consistent trends across food categories grouped by physical characteristics.
For sphere-like foods (sago, red bean, mentos), GRITS eliminates most spillage seen in the vanilla diffusion policy, achieving nearly perfect success. The only exception is mentos, where dense packing still causes occasional spillage.
Regular-shaped foods (marshmallow, fudge, penne) show mixed outcomes: guidance reduces spillage, but their properties still limit success. Marshmallow’s soft, deformable texture makes scooping difficult, and fudge suffers from stickiness and dense packing, while penne is more stable and achieves higher success.
Among irregular items, prune remains challenging since its stickiness can pull neighboring pieces and destabilize the scoop. In contrast, ramen snack benefits from guidance, which stabilizes insertion and rotation despite its rippled geometry, leading to high success. 
Finally, even for challenging categories like mixtures (nuts) and liquids (milk tea), GRITS consistently outperforms the vanilla policy, demonstrating greater reliability.

\begin{figure}[!ht]
    \centering
    \includegraphics[width=0.45\textwidth]{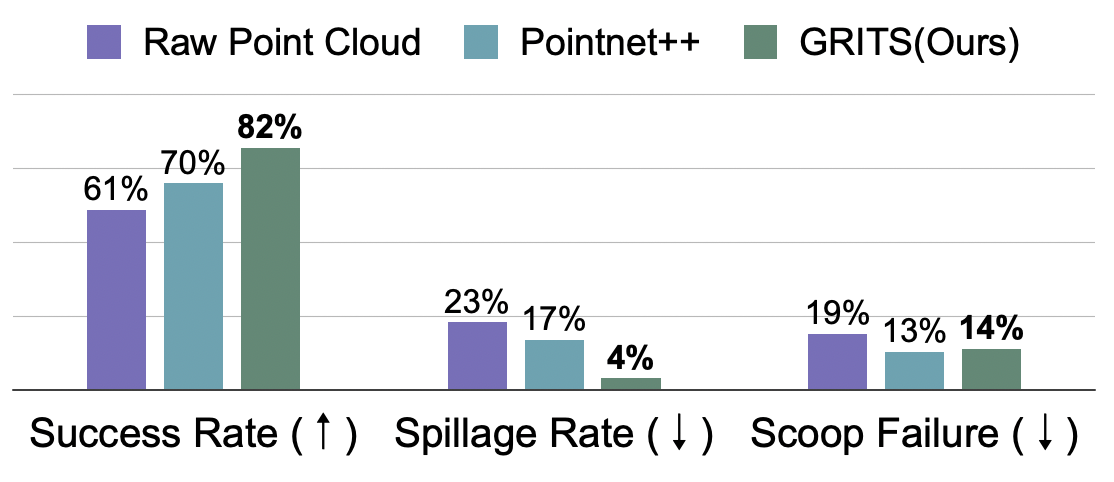}
    \caption{\textbf{Ablation study of spillage predictor.} 
   We compare three variants: (1) raw point cloud input at real-world inference, (2) encoder replaced with original PointNet++~\cite{qi2017pointnet++}, and (3) GRITS method with composite point cloud (food from raw camera data, spoon and bowl from known object models) input and DP3~\cite{ze20243d} encoder.
   The results show that GRITS achieves the highest success rate and lowest spillage by suppressing noise while preserving critical features.
    }
    \label{fig:ablation}
\end{figure}

\subsection{Ablation Study of Spillage Predictor}

We further study the spillage predictor design.
%
To narrow the sim-to-real gap, we use composite point clouds that combine depth-reconstructed food with known spoon and bowl models, reducing point dropouts and geometric distortion.
We validate this choice by comparing against raw sensory point clouds and by replacing the DP3 encoder~\cite{ze20243d} with original PointNet++~\cite{qi2017pointnet++}.
%
%
As shown in Fig.~\ref{fig:ablation}, raw point clouds suffer from noise and missing geometry, while PointNet++ struggles to capture spillage-relevant object and motion patterns.
We often observe that these weaker predictors fail to raise alerts when trajectories approach high-risk states, so corrective refinements are triggered too late, sometimes only after spillage has already occurred.
In contrast, composite point clouds with DP3 enable more effective feature extraction, leading to higher success rates and significantly lower spillage. Overall, both components are crucial for robust spillage-aware guidance.

\section{LIMITATIONS and FUTURE WORK}

Our experiments reveal that failure cases often arise from insufficient information about food physical properties beyond visual cues.
Future work could incorporate pre-interaction strategies~\cite{tai2023scone} and multimodal sensing~\cite{sundaresan2022learning}, including force-torque and tactile feedback, to build a more comprehensive representation of food characteristics.
In addition, the current approach relies on known spoon and bowl models to enhance stabilization. Extending the framework to handle varying settings would improve generalization.
Beyond scooping, the framework may extend to other food manipulation tasks such as cutting and pouring, with task-specific objectives and constraints left for future work.

\section{CONCLUSIONS}

In this work, we present GRITS, a spillage-aware guided diffusion policy framework for robotic food scooping. By integrating a spillage predictor as a guidance mechanism, GRITS generates adaptive trajectories that minimize spillage while maintaining high success rates. 
By leveraging diffusion models, the framework handles diverse food properties and refines trajectories at test time with real-time feedback. 
Real-world experiments show that GRITS consistently outperforms existing methods in both success rate and spillage reduction across a broad range of food categories, including unseen and challenging cases, demonstrating its effectiveness and generalizability.

%
%






\section*{ACKNOWLEDGMENT}

The work is sponsored in part by the National Science and Technology Council under grants 113-2634-F-002-007-, 114-2634-F-A49-004-, the Center for Intelligent Team Robotics and Human-Robot Collaboration under the “Top Research Centers in Taiwan Key Fields Program” of the Ministry of Education, Taiwan, and the Ministry of Education, and the Yushan Fellow Program Administrative Support Grant.



\bibliographystyle{IEEEtran}
\bibliography{root.bib}

\end{document}